# Towards the Automatic Coding of Medical Transcripts to Improve Patient-Centered Communication

[1]Gilchan Park, [1]Julia Taylor Rayz, [2]Cleveland G. Shields
[1]Computer and Information Technology, [2]Human Development & Family Studies, Purdue University
{ park550@purdue.edu jtaylor1@purdue.edu cgshield@purdue.edu }

## ABSTRACT

This paper aims to provide an approach for automatic coding of physician-patient communication transcripts to improve patient-centered communication (PCC). PCC is a central part of high-quality health care. To improve PCC, dialogues between physicians and patients have been recorded and tagged with predefined codes. Trained human coders have manually coded the transcripts. Since it entails huge labor costs and poses possible human errors, automatic coding methods should be considered for efficiency and effectiveness. We adopted three machine learning algorithms (Naïve Bayes, Random Forest, and Support Vector Machine) to categorize lines in transcripts into corresponding codes. The result showed that there is evidence to distinguish the codes, and this is considered to be sufficient for training of human annotators.

## INTRODUCTION

One of the key components in the quality of health care is the measurement of patient care, and measuring clinical practice is a crucial area to be enhanced. In particular, PCC is a fundamental element of high-quality health care. Incomplete or incorrect understanding about treatment between physicians and patients leads to unexpected and unpleasant results for patients. In a patient-centered health care system, effective communication between clinicians and patients is critical (Baker, 2001). Given the fact that the measurement of clinical care quality is mostly done by manually reviewing medical records which entails huge labor cost and inaccurate information, people in medical environment have acknowledged the need for efficient approaches to quality measurement. As a way of improving PCC, dialogues between physicians and patients have been analyzed using prognosis communication codes. There are several reasons to undertake the analysis of communication between doctors and patients. One is to ensure that certain policies are enforced; two is for doctor or medical personnel training; three is to analyze the effectiveness of communication. Today, trained human coders have manually coded the transcripts. Since it is costly and time consuming, automatic coding methods should be considered for efficiency and effectiveness.

## REALTED WORK

### PCC definition & necessity & usages & usefulness & guidelines

PCC is a communication among clinicians, patients, and family members that makes patient-centered care possible. Patient-centeredness is based on moral philosophy with core elements: understanding patients' needs, perspectives, and expectations; assisting patients' participation in understanding their clinical conditions; improving the patient-clinician relationship (Epstein et al., 2005a). Choosing an effective tool to measure PCC is challenging and essential to obtain information about communication behaviors and effects.

To effectively measure the quality of patient care, clinical guidelines have been implemented and distributed. The definition of clinical practice guidelines is "systematically developed statements to assist practitioner and patient decisions about appropriate health care for specific clinical circumstances" (Field & Lohr, 1992). However, clinical guidelines have not been widely used for improving health care. The main reasons include indefinite instructions and formats for guideline development process and the lack of assessment measurements for compliance with guidelines. To render guidelines less ambiguous, Shiffman et al. proposed an approach that enabled clinical practice guideline creators to author their guidelines using controlled vocabulary that restricted grammar and lexicon (Shiffman et al., 2009). Harkema et al. measured and assessed the quality of colonoscopy procedures for colorectal cancer screening based on free text portions in electronic health records that contained salient information for guideline measurements (Harkema et al., 2011). In the smoking-cessation care research, the rule-based clinical event classifier, MediClass, determined whether clinicians conformed to the 5A's of smoking-cessation care with high accuracy (Hazlehurst et al., 2005a, 2005b).

Given the fact that the concept of patient-centered care is not well understood in healthcare community, and policies and guidelines related to patient centered care have not been widely used, it is necessary to create practical and effective policies and guidelines to encourage and assess PCC (Epstein et al., 2010). With the guidelines, clinicians can be trained and make appropriate decisions in certain clinical situations. In particular, when discussing bad news, clinicians should practice and prepare giving bad news; assess patient understanding; present information that is easy to understand; provide emotional support; consider individual preferences (Ngo-Metzger et al., 2008). The study revealed that physicians often asked closed-end questions requiring "yes", "no" to patients, and emphasized the importance of teaching physicians how to ask open-ended inquires that elicit patient's narrative story. One approach to assess PCC is a direct observation of clinical encounters that are recorded and reviewed by trained coders (Roter & Larson, 2002; Stiles & Putnam, 1992; Street & Millay, 2001). Participants' needs, expectations, and feelings can be derived from the analysis of observational data (Epstein et al., 2005b). The outcomes of analysis can be used as foundations for PCC guidelines.

## EXPERIMENTAL SETUP

Our goal is to analyze doctor-patient communication according to Prognosis Communication Manual principles. Previously, transcribed doctor patient records have been manually annotated. This study aims at annotating the transcripts automatically by classifying the lines of transcripts, if appropriate, according to the codes of the manual.

### Prognosis Communication Manual

Prognosis Communication Manual, developed by CGS, is a guideline to encode patient-physician communication transcripts with eleven predefined codes. The specifications of the codes are described with examples in the Table 1.

**Table 1 Prognosis communication manual**

| Code | Specification / Example |
|---|---|
| CancKnowl | Assessing patient's knowledge of state of disease |
| | *"Tell me what you understand about where your cancer is now."* |
| OpenDoor | Asking if the patient wants to know about the prognosis, survival, curability/the future or indicating common questions that people have about the prognosis, survival, curability, future quality of life, or palliative care |
| | *"Would you want to know more about what the future holds for you?"* |
| UnderSProg | Assessing the patient's understanding of his/her prognosis |
| | *"What do you understand about your prognosis?"* |
| ChgforWorse | Discussion of how the disease trajectory is changing for the worse without explicit mention of survival time or curability |
| | *"Especially because your disease has been progressing after two types of chemo."* |
| FurQol | Discussion of quality of life in the future |
| | *"Down the road, you likely won't have as much energy to go out and socialize like you do now."* |
| PallCare | Discussing palliative care treatment |
| | *"Have you had thoughts about considering stopping the chemo and focusing more on comfort and quality of life?"* |
| AdvDirect | Discussing advanced directives |
| | *"Have you and your family discussed issues like life support and power of attorney?"* |
| Curability | Discussing if the cancer can be cured |
| | *"Although the treatments might help you live longer and live better, they cannot completely get rid of the cancer. It's not a cure."* |
| SurvTime | Discussing estimates of survival time |
| | *"I'm not sure that you will make it to Christmas this year."* |
| BestWorstCase | Discussing best case and worst case scenario |
| | *"Most patients will survive 6-12 months. Some may do better and live up to 2 years, and some do worse - only 2-3 months."* |
| DoubFram | Double Framing Survival/Curability Estimates |
| | *"The research shows that about 70% of patients like you will be cured and 30% will relapse."* |

The CancKnowl, OpenDoor, UnderSProg codes are only applicable to physicians' utterances, and the other codes are used for both sides of physicians and patients. The scope of our research is restricted to the portion of physician's codes.

### Dataset

The raw dataset consists of total 839 transcript files, and 492 cases exist within the transcripts. Each case represents a single conversation between a doctor and a patient. The transcript files were coded by human coders. Cases were coded by different number of human coders. To avoid a situation in which a certain portion of text appears more frequent because the case containing the text is coded by many coders than others, the majority opinion of human coders was initially considered. Specifically, when more than half of the coders agree on the particular code for a certain line (single utterance) in a case, then the line will be used for the analysis. However, while applying the majority opinion approach, we found that the level of interrater agreement was significantly low, and this led to the lack of data to be retrieved. Alternatively, we only removed duplicate lines by the same code. For instance, if a certain line is tagged with the same code by multiple coders, duplicates of the line will be discarded. If the line is coded by different codes, then the line will be used in each code category. The rationale of this is that if a line is judged differently by humans, then the line might have potential evidence for each code.

### Data Preprocessing

The data preprocessor first determines whether a line in a transcript is coded or not. Then, it clusters lines with the same code, and stores them in text files. Each code text file is parsed for sentence boundaries. In our approach, we decided to consider nouns, verbs, adjectives, and adverbs rather than take into account all words in sentences. To achieve this, some of natural language techniques were adapted. First, the sentences are tagged with parts of speech using Stanford part of speech tagger (Toutanova et al., 2003). To handle variants of words, words were stemmed using Porter stemmer (Porter, 2001). Stop words were removed as well. Figure 1 illustrates the text preprocessing. Each code file (e.g. AdvDirect.txt, UnderSProg.txt) contains all texts coded by its code. Instance file contains terms of a single utterance.

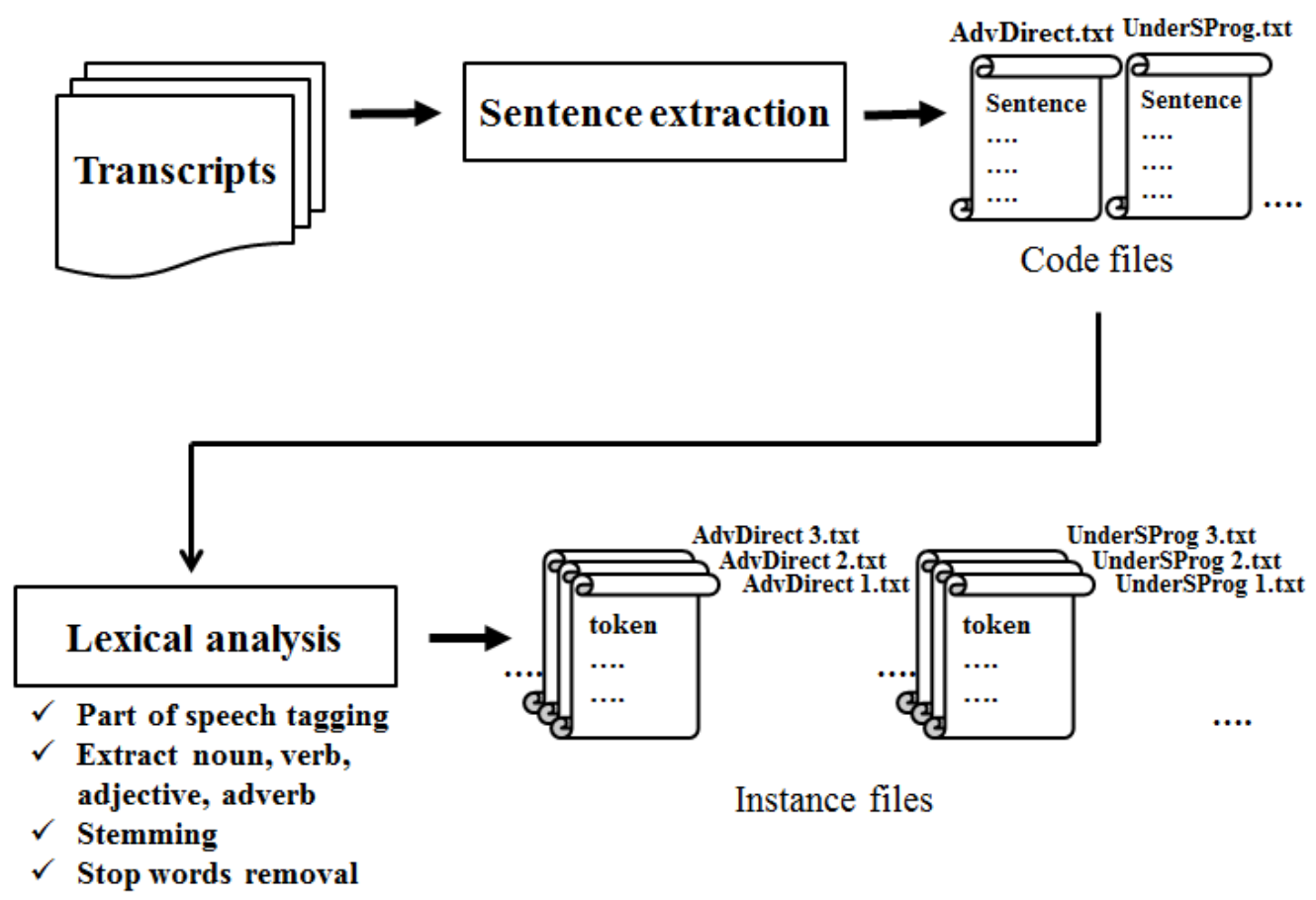

**Fig. 1 Text preprocessing schematic diagram**

### Experiments using Machine Learning Algorithms

To conduct the experiments, we applied three machine learning (ML) techniques: Naïve Bayes (NB), Random Forest (RF), and Support Vector Machine (SVM). The machine learning tool Weka was used to run these algorithms (Hall et al., 2009). The preprocessed dataset revealed that the codes BestWorstCase, CanKnowl, DoubFram, OpenDoor, and UnderSProg were coded too rarely to be compared with the other codes, and therefore were not considered for further processing. The classes used in ML algorithms included AdvDirect, ChgforWorse, Curability, FurQol, PallCare, SurvTime, and NotCoded (a class for lines not coded by any code).

The experiments were largely comprised of three phases. The first experiment evaluated the three ML algorithms' performance for multiclass classification by increasing the minimum number features (words) in each instance (line). The first group contains all instances, the second group contains instances having more than or equal to three words, and the last group contains instances having more than or equal to five words. Since the occurrences among codes were different, the same number of instances for each code was chosen so that the ML classifiers were not over trained by large classes. Table 2 below shows the overall description of data sets.

**Table 2 The number of instances for each code in the first experiment**

| Code | A group (#words ≥1) | B group (#words ≥ 3) | C group (#words ≥5) |
|---|---|---|---|
| AdvDirect | 190 | 190 | 190 |
| ChgforWorse | 190 | 190 | 190 |
| Curability | 190 | 190 | 190 |
| FurQol | 190 | 190 | 190 |
| PallCare | 190 | 190 | 190 |
| SurvTime | 190 | 190 | 190 |
| NotCoded | 190 | 190 | 190 |

For the feature extraction, Chi-square feature selection method was applied. Term frequency, inverse document frequency, and document length normalization were used for term weighting calculation.

Each ML algorithm was performed with 10-fold cross-validation. In each group, ML algorithms were run 4 times with 4 different data sets that were randomly selected from the entire instances for each code. Figure 2, 3, and 4 represent the results of NB, RF, and SVM respectively. Since the classifiers assign exactly one class to each instance, micro-average F1 is the same as accuracy. The results showed that SVM performed better than the other two, and as the number of words increased, both F1 scores slightly became higher. Both results are significantly better than random, as the classification was done between 7 classes resulting in random accuracy of 14%.

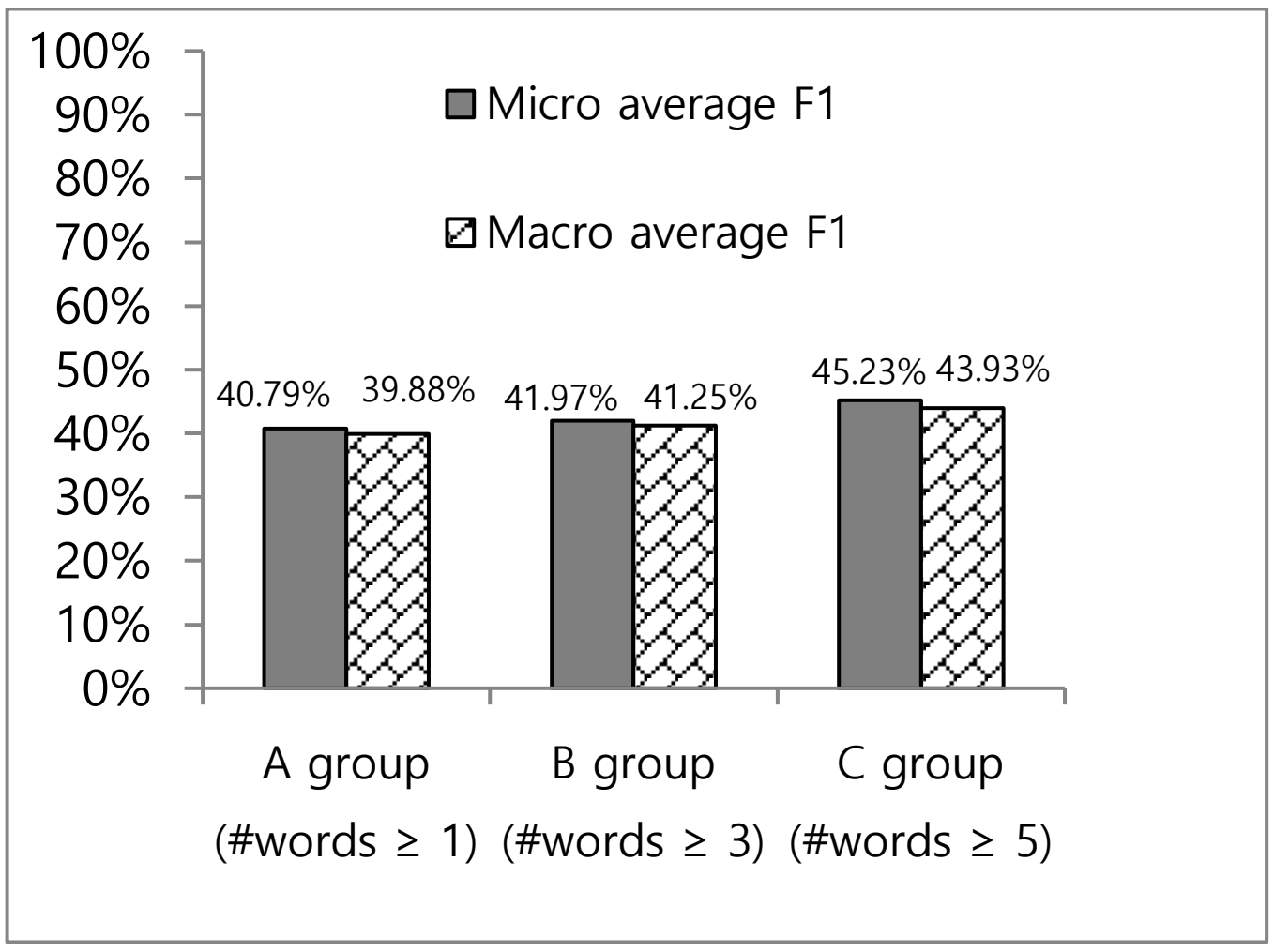

**Fig. 2 Naive Bayes classification result**

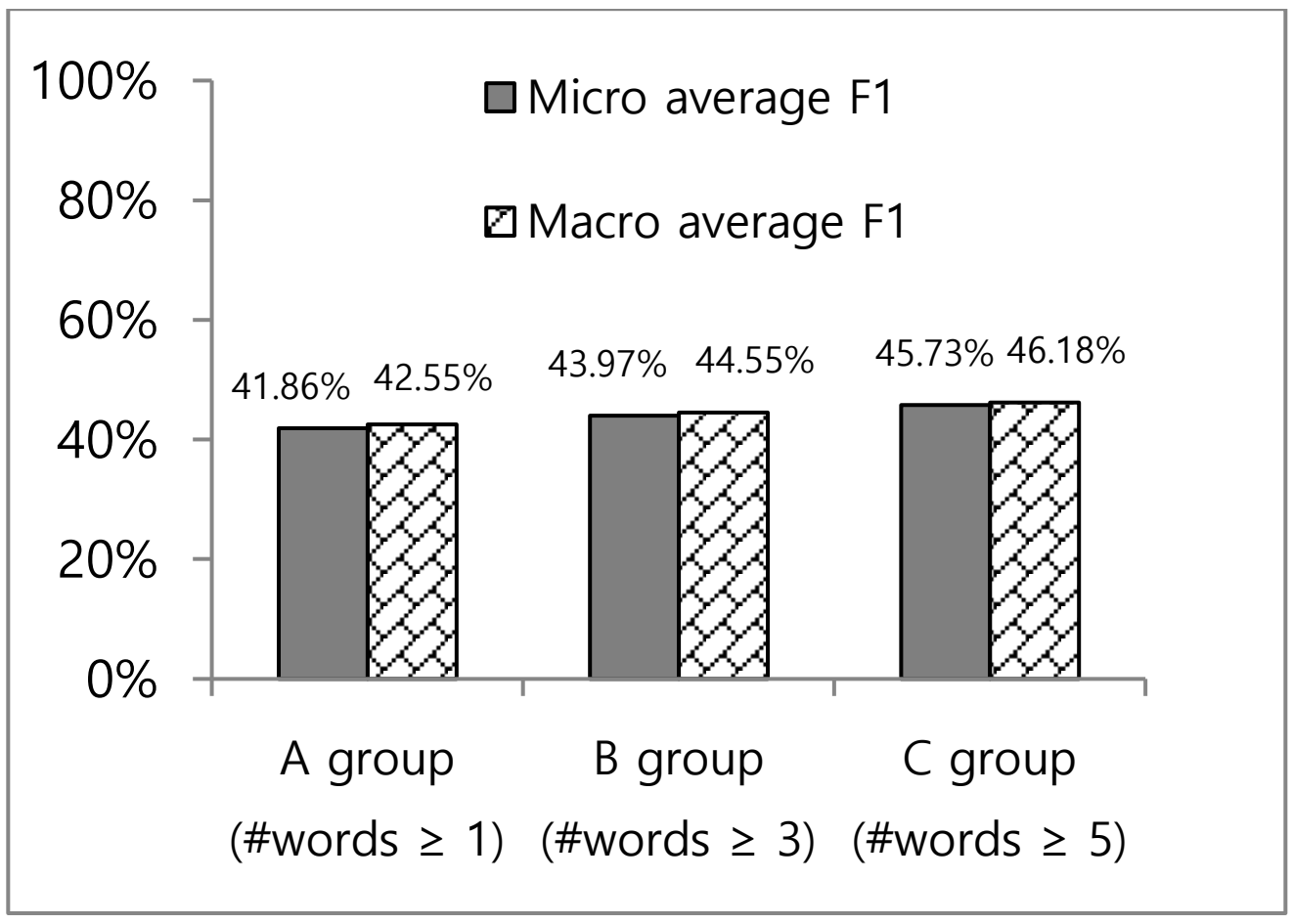

**Fig. 3 Random Forest classification result**

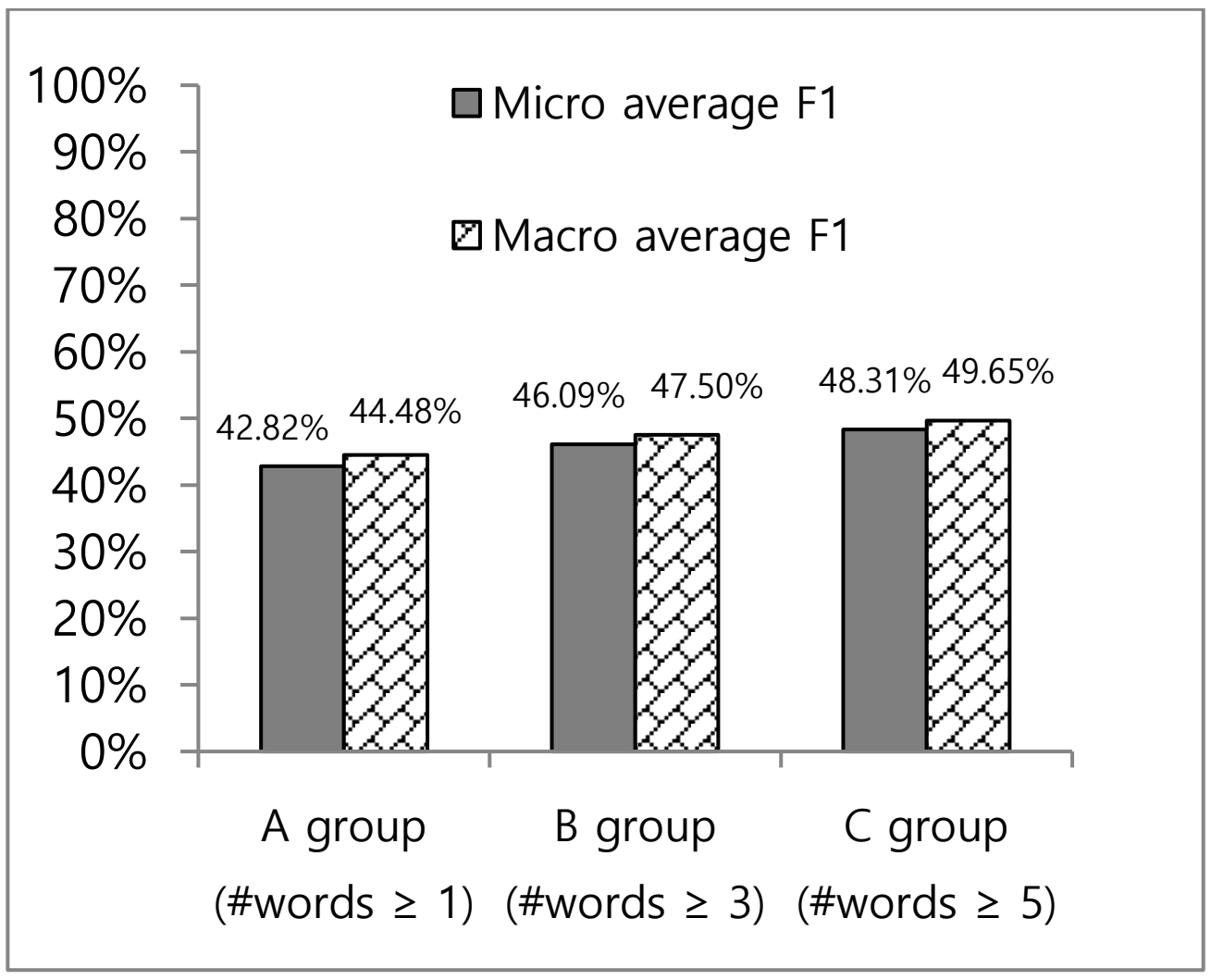

**Fig. 4 SVM classification result**

The second experiment was to check whether the correctly classified instances in C group (#words ≥ 5 words) would be still correctly classified within dataset combined with instances of other groups (#words ≥ 3 and #words ≥ 1). The purpose of this is to determine whether there is a consistent evidence to be distinguishable among codes. In other words, would a line of more than 5 words that was correctly classified be also correctly classified in a more-than-or-equal-to 3 classifier and more-than-or-equal-to 1 classifier? Are specific longer lines better classified with different classifiers? To accomplish this, the data set for A group was reconstructed with the combination of correctly classified instances in C group and the randomly selected portions of samples in A group for the rest. The same applied to the data set for B group. Figure 5 depicts an example of the process of data set construction.

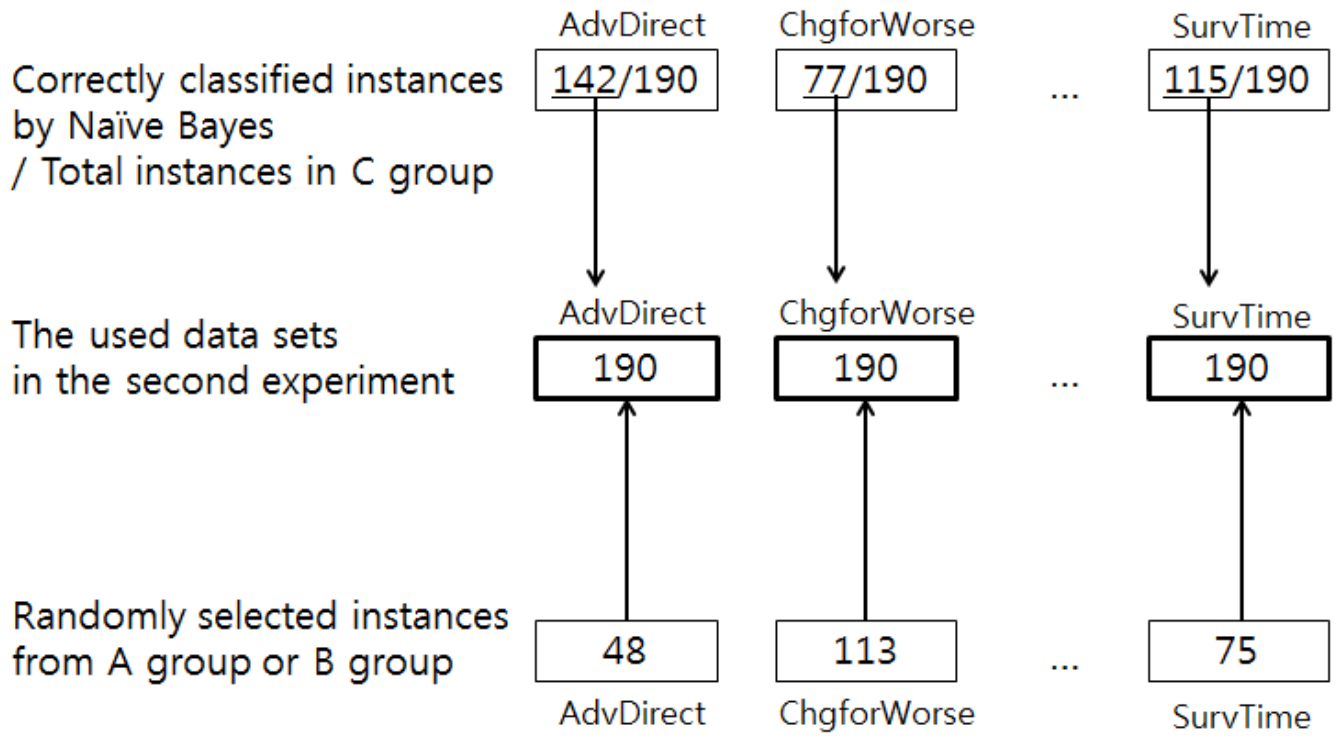

**Fig. 3 The example of construction of data sets in the second experiment**

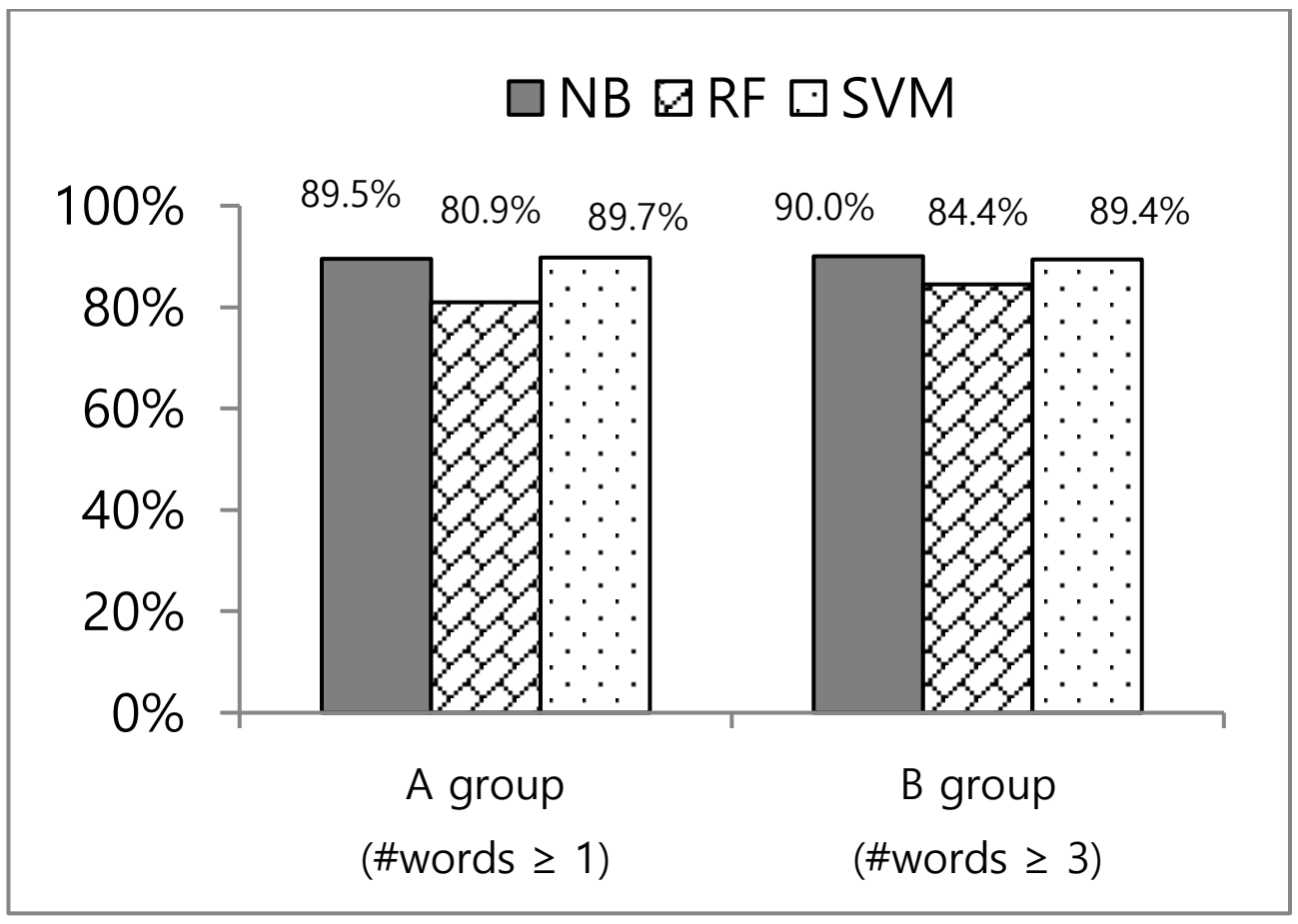

**Fig. 4 Accuracy of correct classification on the instances that were correctly classified in C group in A group and B group**

The combined data sets were classified by NB, SVM, and RF algorithms. The correctly classified instances in C group were tracked to identify whether they were still correctly classified in different groups. Figure 6 describes the accuracy of correct classification on the instances that were correctly classified in C group. The result showed that most of correctly classified instances in C group were still correctly classified in A group and B group.

The third experiment took into account immediately preceding sentences from another dialog participant that led to a coded line response. For instance, the single line *"D: Subsequent tumor markers have come back more elevated than before."* is coded as *'ChgforWorse'*. For the line with previous lines include its preceding successive lines by the same person: *"D: Okay, so we have a lot to talk about today and your treatment, I think is starting today as well. D: I did the orders and tried to get it started as soon as possible. D: I had an inkling obviously that things were starting to grow between your symptoms and the tumor markers. D: Subsequent tumor markers have come back more elevated than before."* Our assumption of this is that previous lines can be strongly related to the coded line. In other words, the previous lines can have features for machines to detect the coded line.

Two sets were prepared for the experiment: single coded lines and single coded lines with previous lines. The preceding experiments were all conducted using single coded lines. The instances in both sets contained 5 words or more. This time, *'NotCoded'* class was excluded since we assumed that the lines in the class would not be closely related unlike other codes. The data sets consist of 6 classes with 190 instances for each class. The NB, RF, and SVM algorithms classified the data. The ML algorithms were run 4 times with 4 different data sets that were randomly selected from the entire instances for each code. Figures 7, 8, 9 shows the results of each algorithm using 10-fold cross validation.

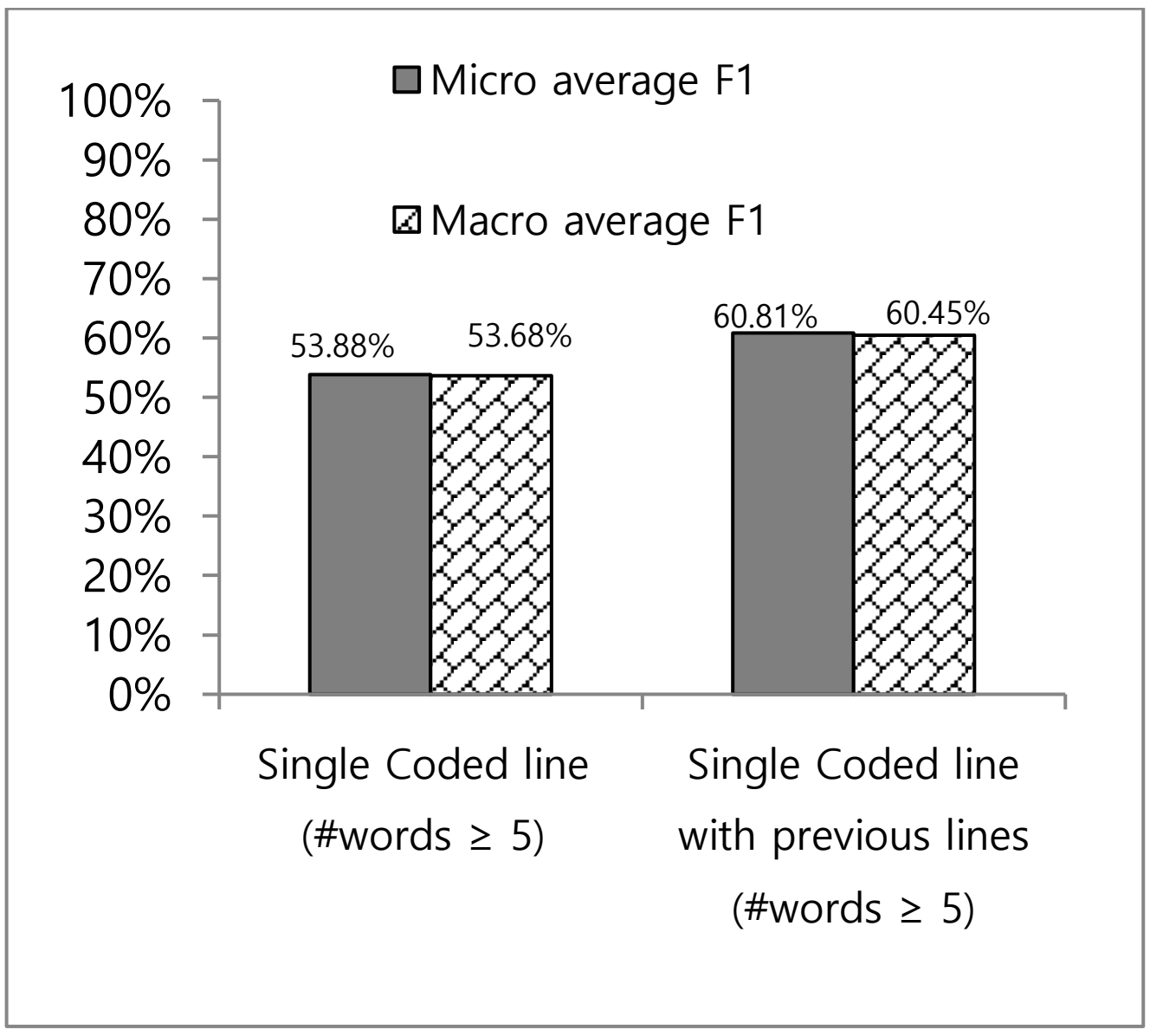

**Fig. 7 Naive Bayes classification result**

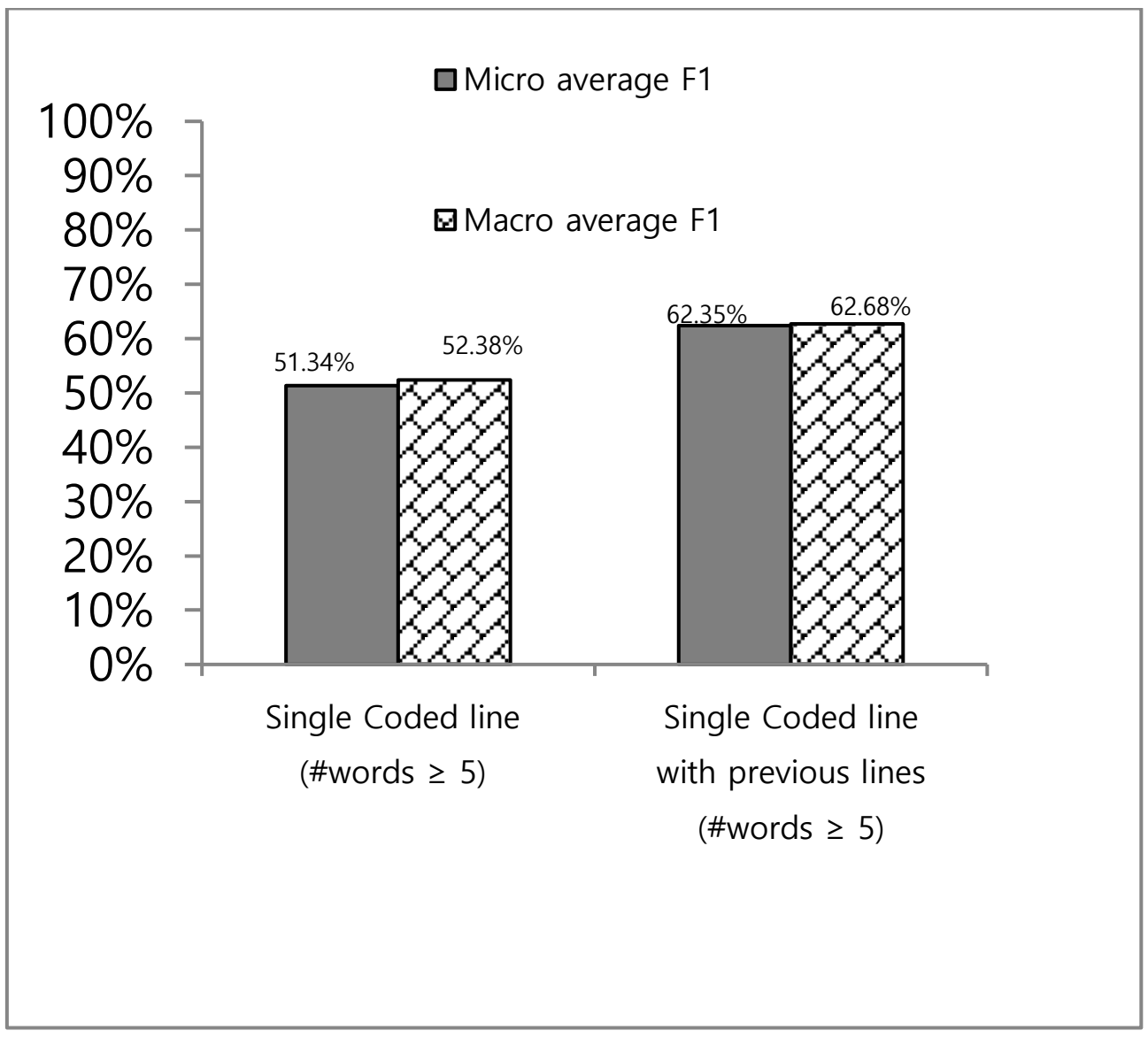

**Fig. 8 Random Forest classification result**

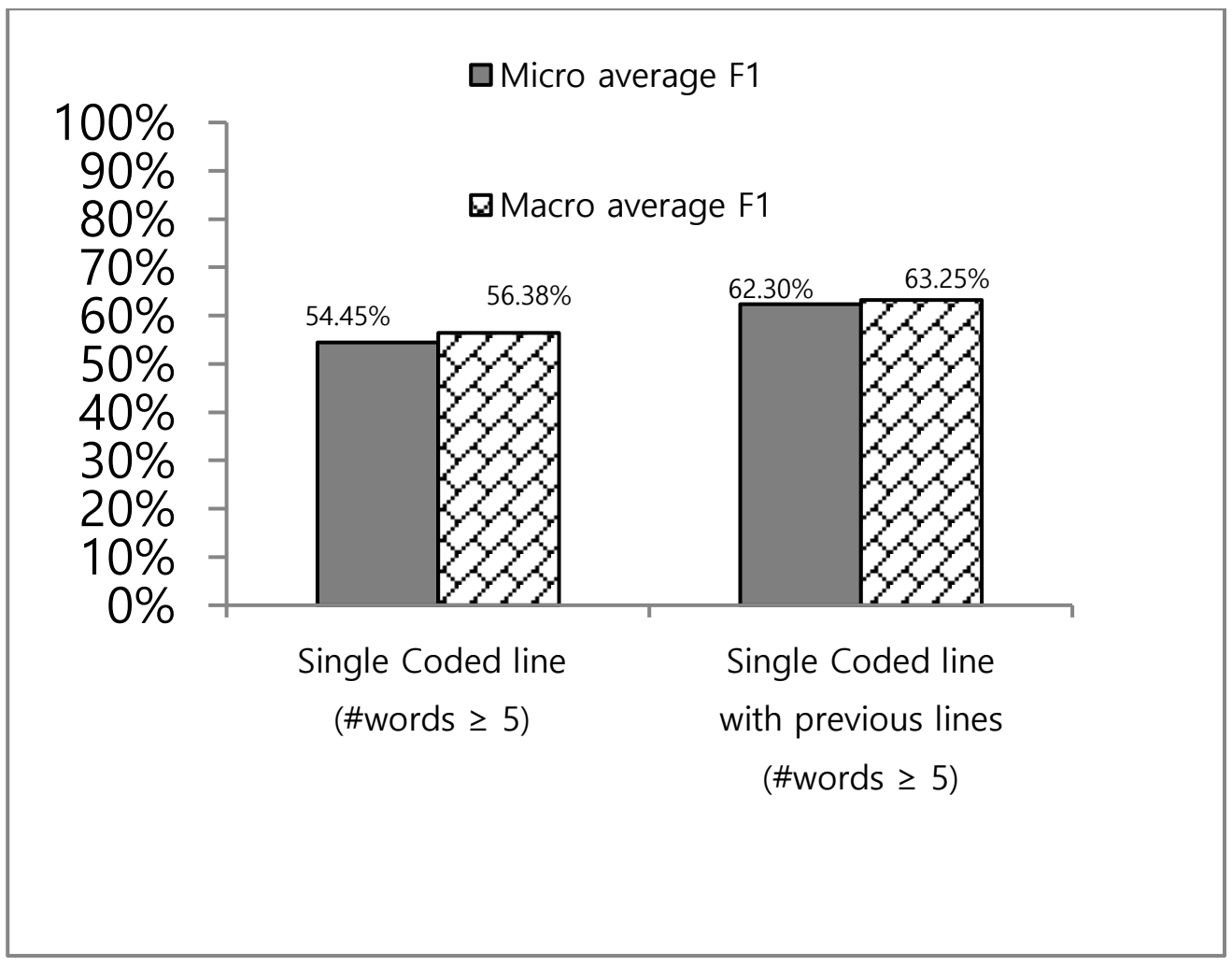

**Fig. 9 SVM classification result**

The results indicated that ML algorithms were able to more accurately classify the coded lines with previous lines than the single coded lines with 6 ~ 10% higher accuracy.

## CONCLUSIONS

This paper suggested an approach for automatic coding of physician-patient communication transcripts to improve PCC. The experiments were conducted in two phases. The first experiment was to evaluate NB, RF, SVM algorithms' performance on code classification. The results indicated that SVM performed better than the other two. In the results, we discovered a couple of causes that affected the performance. One reason can be the length of each sample. Each sample is a single utterance of a physician's speech. After the preprocessing step, the majority of samples contained a single word. In other words, this might lead to the lack of evidence for classifiers to distinguish samples. This was proven by the fact that the classification accuracy became greater when the minimum number of words in data sets increased. The difference in accuracy between A group (#words ≥ 1) and C group (#words ≥ 5) was around 5%. The low interrater agreement can affect the low performances. Human coders coded the transcripts differently, and thereby a certain line (utterance, sample in ML) belongs to more than one category. While this led to ambiguous samples for classifiers, this lack of agreement is one of the reasons for employing machine-classification that removes human bias (although introduces many other issues). The majority opinion was not applicable in this research since it considerably reduced the size of usable samples. With a larger number of transcripts that are coded by many people the majority opinion could be considered, and the performance is expected to improve. While the results are significantly better than random, they are not reliable enough to be used in real life. One of the reasons is the small sample size that was tagged. In order not to over-train classifiers by large classes, the used samples for each code were restricted to 190; however, this sample size might not be sufficient to train classifiers. The second experiment was to identify consistent samples in data sets. The correctly classified samples in C group (#words ≥ 5) were still correctly classified in different data sets with 80% ~ 90% accuracy. This indicates that there exist distinguishable samples between codes. The proved samples can be used for feature selection to calculate term weightings, or can be representative data for other approaches such as concept identification in semantic processing. The third experiment was set to measure relatedness between coded lines with their preceding lines. The results showed that the machine more correctly classified the coded lines in combination with preceding lines than the single coded lines with 6 ~ 10% higher accuracy. This proved our assumption that utterances in a specific time can belong to the same or at least similar context. In the experiment, we excluded 'NotCoded' class because the utterances in the class would not be close due to varied contexts. The level of consistency in 'NotCoded' class will be further investigated.

In this research, not all codes were considered due to the lack of samples for some codes. When more transcript data that contain enough samples for the currently discarded codes are given, all codes will be analyzed and the possible reasons and causes of low performance mentioned above will be further proved and addressed. The analysis of texts was restricted to physicians' portion of transcripts. Since patient-centeredness must consider characteristics of patients, a comprehensive measure of PCC is required, and we will achieve this in our future work. However, even with the result of classification between 7 classes being 50%, this is enough to be used for training of people that manually tag the data today, thus making a step forward in producing better results in PCC in general and prognosis communication manual-based tagging in particular.

## ACKNOWLEDGMENT

This work was supported by grants from the National Cancer Institute (R01 CA140419-05; co-PIs: RM Epstein & RL Kravitz, R01CA168387; co-PIs: PR Duberstein & H Prigerson, and R01-CA155376; co-PIs CG Shields & RM Epstein).